\documentclass[11pt]{article}

\usepackage{acl2012}
\usepackage{times}
\usepackage{latexsym}
\usepackage{amsmath}
\usepackage{multirow}
\usepackage{url}
\usepackage{color} 

\usepackage{graphicx}

\usepackage{algorithm}
\usepackage{algpseudocode}

\usepackage{stmaryrd}
\usepackage{amsfonts}
\usepackage{amssymb}

\title{A Hierarchical Distance-dependent Bayesian Model for \\ Event Coreference Resolution}

\author{
Bishan Yang \;\;\;\;\;\; Claire Cardie\\
Department of Computer Science\\
Cornell University\\
{\tt \{bishan, cardie\}@cs.cornell.edu}\\
\And
Peter Frazier\\
      School of Operations Research \\
      and Information Engineering\\ 
Cornell University\\
{\tt pf98@cornell.edu}\\
}

\date{}

\begin{document}
\maketitle
\begin{abstract}
We present a novel hierarchical distance-dependent Bayesian model for event coreference resolution. While existing generative models for event coreference resolution are completely unsupervised, our model allows for the incorporation of pairwise distances between event mentions ---  information that is widely used in supervised coreference models to guide the generative clustering processing for better event clustering both within and across documents. We model the distances between event mentions using a feature-rich learnable distance function and encode them as Bayesian priors for nonparametric clustering. Experiments on the ECB+ corpus show that our model outperforms state-of-the-art methods for both within- and cross-document event coreference resolution. 
\end{abstract}

\section{Introduction}
The task of {\it  event coreference resolution} consists of identifying text snippets that describe events, and then clustering them such that all {\it event mentions} in the same partition refer to the same unique
event. Event coreference resolution can be applied within a single document or across multiple documents and is crucial for many natural language processing tasks including
topic detection and tracking, information extraction, question answering and textual entailment~\cite{bejan2010unsupervised}.  More importantly, event coreference resolution is a necessary component
in any reasonable, broadly applicable computational model of natural language understanding~\cite{humphreys1997event}.

\begin{figure}[ht]
\centering
\includegraphics[width=1.0\linewidth]{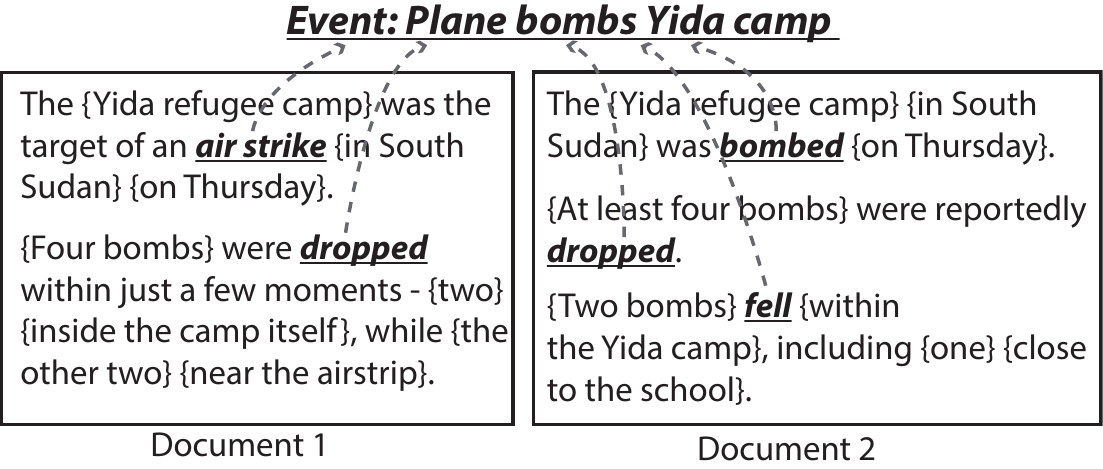}
\label{coref_example}
\caption{Examples of event coreference. Mutually coreferent event mentions are underlined and in boldface; participant and spatio-temporal information for the highlighted event is marked by curly brackets.}
\end{figure}

In comparison to entity coreference 
resolution~\cite{ng2010supervised}, which 
deals with identifying and grouping noun phrases that
refer to the same discourse entity, event coreference resolution has not been extensively studied. This is, in part, because events typically exhibit a more complex structure than entities: a single event can be described via multiple event mentions, and
a single event mention can be associated with multiple {\it event arguments} that characterize the participants in the event as well as spatio-temporal information~\cite{bejan2010unsupervised}. Hence, the coreference decisions for event mentions usually require the interpretation of event mentions and their arguments in context. See, for example, Figure~\ref{coref_example}, in which 
five event mentions across two documents all refer to the same underlying event: \textit{Plane bombs Yida camp}.

Most previous approaches to event coreference resolution (e.g.,~\newcite{ahn2006stages}, \newcite{chen2009pairwise}) operated by extending the supervised pairwise classification model that is widely used in entity coreference resolution (e.g.,\ \newcite{ng2002improving}).  In this framework, pairwise distances between event mentions are modeled via event-related features (e.g.,\ that indicate event argument compatibility), and agglomerative clustering is applied to greedily merge event mentions into clusters. A major drawback of this general approach is that it makes hard decisions on the merging and splitting of clusters based on heuristics derived from the pairwise distances. In addition, it only captures pairwise coreference decisions within a single document and can not account for signals that commonly appear across documents. More recently, Bejan and Harabagiu~\shortcite{bejan2010unsupervised,bejan2014unsupervised} proposed several nonparametric Bayesian models for event coreference resolution that probabilistically infer event clusters both within a document and across multiple documents. Their method, however, is completely unsupervised, and thus can not encode any readily available supervisory information to guide the model toward better event clustering.

To address these limitations, we propose a novel Bayesian model for within- and cross-document event coreference resolution. It leverages supervised feature-rich modeling of pairwise coreference relations and generative modeling of cluster distributions, and thus allows for both probabilistic inference over event clusters and easy incorporation of pairwise linking preferences. Our model builds on the framework of the distance-dependent Chinese restaurant process (DDCRP)~\cite{blei2011distance}, which was introduced to incorporate data dependencies into nonparametric clustering models. 
Here, however, we extend the DDCRP to allow the incorporation of feature-based, learnable distance functions as clustering priors, thus encouraging event mentions that are close in meaning to belong to the same cluster. In addition, we introduce to the DDCRP a representational hierarchy that allows event mentions to be grouped within a document and within-document event clusters to be grouped across documents. 

To investigate the effectiveness of our approach, we conduct extensive experiments on the ECB+ corpus~\cite{cybulska2014using}, an extension to EventCorefBank (ECB)~\cite{bejan2010unsupervised}
and the largest corpus available that contains event coreference annotations within and across documents. We show that integrating pairwise learning of event coreference relations with unsupervised hierarchical modeling of event clustering achieves promising improvements over state-of-the-art approaches for within- and cross-document event coreference resolution.

\section{Related Work}
Coreference resolution in general is a difficult natural language processing (NLP) task and typically requires sophisticated inferentially-based knowledge-intensive models~\cite{kehler2002coherence}. Extensive work in the literature focuses on the problem of entity coreference resolution and many techniques have been developed, including rule-based deterministic models (e.g.~\newcite{cardie1999noun}, \newcite{raghunathan2010multi}, \newcite{lee2011stanford}) that traverse over mentions in certain orderings and make deterministic coreference decisions based on all available information at the time; supervised learning-based models (e.g.~\newcite{stoyanov2009conundrums}, \newcite{rahman2011coreference}, \newcite{durrett2013easy}) that make use of rich linguistic features and the annotated corpora to learn more powerful coreference functions; and finally, unsupervised models (e.g.~\newcite{bhattacharya2006latent}, Haghighi and Klein (2007, 2010)) \nocite{haghighi2007unsupervised} \nocite{haghighi2010coreference} that successfully apply generative modeling to the coreference resolution problem.  

Event coreference resolution is a more complex task than entity coreference resolution~\cite{humphreys1997event} and also has been relatively less studied. Existing work has adapted similar ideas to those used in entity coreference.~\newcite{humphreys1997event} first proposed a deterministic clustering mechanism to group event mentions of pre-specified types based on hard constraints. Later approaches~\cite{ahn2006stages,chen2009pairwise} applied learning-based pairwise classification decisions using event-specific features to infer event clustering. Bejan and Harabagiu~\shortcite{bejan2010unsupervised,bejan2014unsupervised} proposed several unsupervised generative models for event mention clustering based on the hierarchical Dirichlet process (HDP)~\cite{teh2006hierarchical}. Our approach is related to both supervised clustering and generative clustering approaches. It is a nonparametric Bayesian model in nature but encodes rich linguistic features in clustering priors. More recent work modeled both entity and event information in event coreference.~\newcite{lee2012joint} showed that iteratively merging entity and event clusters can boost the clustering performance.~\newcite{liu2014supervised} demonstrated the benefits of propagating information between event arguments and event mentions during a post-processing step. Other work modeled event coreference as a predicate argument alignment problem between pairs of sentences, and trained classifiers for making alignment decisions~\cite{roth2012aligning,wolfepredicate}. Our model also leverages event argument information into the decisions of event coreference but incorporates it into Bayesian clustering priors.

Most existing coreference models, both for events and entities, focus on solving the within-document coreference problem. Cross-document coreference has attracted less attention due to lack of annotated corpora and the requirement for larger model capacity. Hierarchical models~\cite{singh2010distantly,wick2012discriminative,haghighi2007unsupervised} have been popular choices for cross-document coreference as they can capture coreference at multiple levels of granularities. Our model is also hierarchical, capturing both within- and cross-document coreference.

Our model is also closely related to the distance-dependent Chinese Restaurant Process (DDCRP)~\cite{blei2011distance}. The DDCRP is an infinite clustering model that can account for data dependencies~\cite{ghosh2011spatial,socher2011spectral}. But it is a flat clustering model and thus cannot capture hierarchical structure that usually exists in large data collections. Very little work has explored the use of DDCRP in hierarchical clustering models. \newcite{kim2011accounting,ghosh2011spatial} combined a DDCRP with a standard CRP in a two-level hierarchy analogous to the HDP with restricted distance functions. \newcite{ghoshnonparametric} proposed a two-level DDCRP with data-dependent distance-based priors at both levels. Our model is also a two-level DDCRP model but differs in that its distance function is learned using a feature-rich log-linear model. We also derive an effective Gibbs sampler for posterior inference.

\section{Problem Formulation}
\label{background}
We adopt the terminology from ECB+~\cite{cybulska2014using}, a corpus that extends the widely used EventCorefBank (ECB \cite{bejan2010unsupervised}). An \textbf{event} is something that happens or a situation that occurs~\cite{cybulska2014guidelines}. It consists of four components: (1) an \textit{Action}: what happens in the event; (2) \textit{Participants}: who or what is involved; (3) a \textit{Time}: when the event happens; and (4) a \textit{Location}: where the event happens. We assume that each document in the corpus consists of a set of mentions --- text spans --- that describe event actions, their participants, times, and locations. Table~\ref{event_rep} shows examples of these in the sentence ``Sudan bombs Yida refugee camp in South Sudan on Thursday, Nov 10th, 2011."
\begin{table}
\begin{footnotesize}
\begin{center}
\begin{tabular}{|c|c|}
\hline
Action & \textit{bombs}\\
\hline
Participant & \textit{Sudan, Yida refugee camp}\\
\hline
Time & \textit{Thursday, Nov 10, 2011}\\
\hline
Location & \textit{South Sudan}\\
\hline
\end{tabular}
\end{center}
\caption{\label{event_rep}Mentions of event components}
\end{footnotesize}
\end{table}

In this paper, we also use the term \textbf{event mention} to refer to the mention of an event action, and \textbf{event arguments} to refer
collectively to mentions of the participants, times and locations involved in the event. Event mentions are usually noun phrases or verb phrases that clearly describe events. Two event mentions are considered \textbf{coreferent} if they refer to the same actual event, i.e.\ a situation involving a particular combination of action, participants, time and location. Note that in text, not all event arguments are always present for an event mention; they may even be distributed over different sentences. Thus whether two event mentions are coreferential should be determined based on the context. For example, in Figure~\ref{coref_example}, the event mention \textit{dropped} in {\sc Document 1} corefers with \textit{air strike} in the same document as they describe the same event, \textit{Plane bombs Yida camp}, in the discourse context; it also corefers with \textit{dropped} in {\sc Document 2} based on the contexts of both documents.

The problem of event coreference resolution can be divided into two sub-problems: (1) \textbf{event extraction}: extracting event mentions and event arguments, and (2) \textbf{event clustering}: grouping event mentions into clusters according to their coreference relations. We consider both within- and cross-document event coreference resolution  and hypothesize that leveraging context information from multiple documents will improve both within- and cross-document coreference resolution. In the following, we first describe the event extraction step and then focus on the event clustering step. 

\section{Event Extraction}
\label{event_extract}
The goal of event extraction is to extract from a text all event mentions (actions) and event arguments (the associated participants, times and locations). One might expect that event actions could be extracted reasonably well by identifying verb groups; and event arguments, by applying semantic role labeling (SRL)
to identify, for example, the \textit{Agent} and \textit{Patient} of each predicate.
Unfortunately, most SRL systems only handle verbal predicates and so
would miss event mentions described via noun phrases. In addition,
SRL systems are not designed to capture event-specific arguments. Accordingly, we found that a state-of-the-art SRL system (SwiRL~\cite{surdeanu2007combination}) extracted only 56\% of
the actions, 76\% of participants, 65\% of times and 13\% of  locations for events in a development set of ECB+ based on a head word matching evaluation measure. (We provide dataset details in Section~\ref{exp}.)

To produce higher recall, we adopt a supervised approach and train an event extractor using sentences from ECB+, which are annotated for
event actions, participants, times and locations. Because these mentions vary widely in their length and grammatical type, we employ semi-Markov CRFs~\cite{sarawagi2004semi}
using the loss-augmented objective of~\newcite{yang2014joint} that
provides more accurate detection of mention boundaries.
We make use of a rich feature set that includes word-level features such as unigrams, bigrams, POS tags, WordNet hypernyms, synonyms and FrameNet semantic roles, and phrase-level features such as phrasal syntax (e.g.,\ NP, VP) and phrasal embeddings (constructed by averaging word embeddings produced by word2vec~\cite{mikolov2013efficient}). Our experiments on the same (held-out) development data show that the semi-CRF-based extractor correctly identifies 95\% of actions, 90\% of participants, 94\% of times and 74\% of locations again based on head word matching.

Note that the semi-CRF extractor identifies event mentions and event 
arguments but not relationships among them, i.e.\ it does not associate arguments with an event mention. Lacking supervisory data in the ECB+ corpus for training an event action-argument relation detector, we assume that all event arguments identified by the semi-CRF extractor are related to all event mentions in the same sentence and then apply SRL-based heuristics to augment and further disambiguate intra-sentential action-argument relations (using the SwiRL SRL). More specifically, we link each verbal event mention to the participants that match its \textit{ARG0}, \textit{ARG1} or \textit{ARG2} semantic role fillers; similarly, we associate with the event mention the time and locations that match its \textit{AM-TMP} and \textit{AM-LOC} role fillers, respectively. For each nominal event mention, we associate those participants that match the possessor of the mention since these were suggested in~\newcite{lee2012joint} as playing the \textit{ARG0} role for
nominal predicates.

\section{Event Clustering}
\newcommand{\pfedit}[1]{{\color{blue} #1}}
Now we describe our proposed Bayesian model for event clustering. Our model is a hierarchical extension of the distance-dependent Chinese Restaurant Process (DDCRP). It first groups event mentions within a document to form within-document event cluster and then groups these event clusters across documents to form global clusters. The model can account for the similarity between event mentions during the clustering process, putting a bias toward clusters comprised of event mentions that are similar to each other based on the context. To capture event similarity, we use a log-linear model with rich syntactic and semantic features, and learn the feature weights using gold-standard data. 

\subsection{Distance-dependent Chinese Restaurant Process}
The Distance-dependent Chinese Restaurant Process (DDCRP) is a generalization of the Chinese Restaurant process (CRP) that models distributions over partitions. In a CRP, the generative process can be described by imagining data points as customers in a restaurant and the partitioning of data as tables at which the customers sit. The process randomly samples the table assignment for each customer sequentially: the probability of a customer sitting at an existing table is proportional to the number of customers already sitting at that table and the probability of sitting at a new table is proportional to a scaling parameter. For each customer sitting at the same table, an observation can be drawn from a distribution determined by the parameter associated with that table. Despite the sequential sampling process, the CRP makes the assumption of exchangeability: the permutation of the customer ordering does not change the probability of the partitions.

The exchangeability assumption may not be reasonable for clustering data that has clear inter-dependencies. The DDCRP allows the incorporation of data dependencies in infinite clustering, encouraging data points that are closer to each other to be grouped together. In the generative process, instead of directly sampling a table assignment for each customer, it samples a customer link, linking the customer to another customer or itself. The clustering can be uniquely constructed once the customer links are determined for all customers: two customers belong to the same cluster if and only if one can reach the other by traversing the customer links (treating these links as undirected).

More formally, consider a sequence of customers $1,...,n$, and denote $\mathbf{a}=(a_1,...,a_n)$ as the assignments of the customer links. $a_i \in \{1,\ldots,n\}$ is drawn from
\begin{equation}
\label{DDCRP-prior}
    p(a_i=j|F,\alpha)\propto 
\begin{cases}
    F(i,j), & j \neq i\\
    \alpha, & j = i\\
\end{cases}
\end{equation}
where $F$ is a distance function and $F(i,j)$ is a value that measures the distance between customer $i$ and $j$. $\alpha$ is a scaling parameter, measuring self-affinity. For each customer, the observation is generated by the per-table parameters as in the CRP. A DDCRP is said to be {\it sequential} if $F(i,j)=0$ when $i<j$, so customers may link only to themselves, and to previous customers.

\subsection{A Hierarchical Extension of the DDCRP}
\label{sec-hddcrp}
We can model within-document coreference resolution using a sequential DDCRP. Imagining customers as event mentions and the restaurant as a document, each mention can either refer to an antecedent mention in the document or no other mentions, starting the description of a new event. However, the coreference relations may also exist across documents --- the same event may be described in multiple documents. Thus it is ideal to have a two-level clustering model that can group event mentions within a document and further group them across documents. Therefore we propose a hierarchical extension of the DDCRP (HDDCRP) that employs a DDCRP twice: the first-level DDCRP links mentions based on within-document distances and the-second level DDCRP links the within-document clusters based on cross-document distances, forming larger clusters in the corpus. 

\begin{figure}[ht]
\centering
\includegraphics[width=0.8\linewidth]{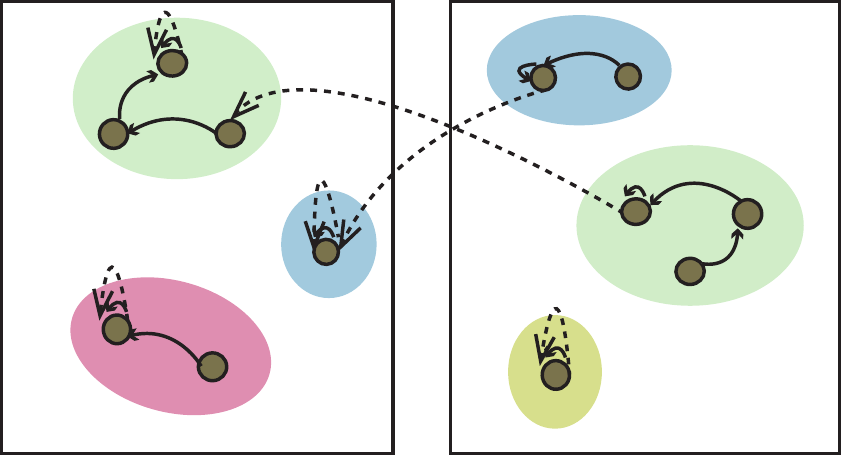}
\caption{\label{fig:HDDCRP_example}\small A cluster configuration generated by the HDDCRP. Each restaurant is represented by a rectangle. The small green circles represent customers. The ovals represent tables and the colors reflect the clustering. Each customer is assigned a customer link (a solid arrow), linking to itself or another customer in the same restaurant. The customer who first sits at the table is assigned a table link (a dashed arrow), linking to itself or another customer in a different restaurant, resulting in the linking of two tables.}
\end{figure}

The generative process of an HDDCRP can be described using the same ``Chinese Restaurant" metaphor. Imagine a collection of documents as a collection of restaurants, and the event mentions in each document as customers entering a restaurant. The local (within-document) event clusters correspond to \textit{tables}. The global (within-corpus) event clusters correspond to \textit{menus} (tables that serve the same menu belong to the same cluster). The hidden variables are the customer links and the table links. Figure \ref{fig:HDDCRP_example} shows a configuration of these variables and the corresponding clustering structure. 

More formally, the generative process for the HDDCRP can be described as follows:
\begin{enumerate}
\item For each restaurant $d\in \{1,...,D\}$, for each customer $i\in \{1,...,n_d\}$, sample a customer link using a sequential DDCRP: 
\begin{equation}
\label{HDDCRP-prior-1}
    p(a_{i,d}=(j,d))\propto 
\begin{cases}
    F_d(i,j), & j < i\\
    \alpha_d, & j = i\\
    0, & j > i
\end{cases}
\end{equation}
\item \label{step:table-link} For each restaurant $d\in \{1,...,D\}$, for each table $t$, sample a table link for the customer $(i,d)$ who first sits at $t$ using a DDCRP:
\begin{equation}
\label{HDDCRP-prior-2}
\begin{split}
&p(c_{i,d}=(j,d'))\propto \\
&\begin{cases}
F_0((i,d),(j,d')), & j\in\{1,...,n_{d'}\}, d'\neq d\\
\alpha_0, & j = i, d' = d\\
\end{cases}
\end{split}
\end{equation}
\item 
\label{step:cluster} 
Calculate clusters $\mathbf{z}(\mathbf{a},\mathbf{c})$ by traversing all the customer links $\mathbf{a}$ and the table links $\mathbf{c}$. Two customers are in the same cluster if and only if there is a path from one to the other along the links, where we treat both table and customer links as undirected.

\item For each cluster $k\in \mathbf{z}(\mathbf{a},\mathbf{c})$, sample parameters $\phi_k\sim G_0(\lambda)$.
\item For each customer $i$ in cluster $k$, sample an observation $x_i\sim p(\cdot|\phi_{z_i})$ where $z_i=k$.
\end{enumerate}

$F_{1:D}$ and $F_0$ are distance functions that map a pair of customers to a distance value. We will discuss them in detail in Section \ref{dist-param}.

\subsection{Posterior Inference with Gibbs Sampling}
\label{posterior}
The central computation problem for the HDDCRP model is posterior inference --- computing the conditional distribution of the hidden variables given the observations $p(\mathbf{a}, \mathbf{c}|\mathbf{x}, \alpha_0, F_0, \alpha_{1:D}, F_{1:D})$. The posterior is intractable due to a combinatorial number of possible link configurations. Thus we approximate the posterior using Markov Chain Monte Carlo (MCMC) sampling, and specifically using a Gibbs sampler. 

In developing this Gibbs sampler, we first observe that the generative process is equivalent to one that, in step~\ref{step:table-link} samples a table link for {\it all} customers, and then in step~\ref{step:cluster}, when calculating $\mathbf{z}(\mathbf{a},\mathbf{c})$, includes only those table links $c_{i,d}$ originating at customers $(i,d)$ that started a new table, i.e.\ that chose $a_{i,d} = (i,d)$.

The Gibbs sampler for the HDDCRP iteratively samples a customer link for each customer $(i,d)$ from 
\begin{equation}
\label{cond-a}
p(a_{i,d}^*|\mathbf{a}_{-(i,d)}, \mathbf{c},\mathbf{x}, \lambda)\propto p(a_{i,d}^*)H_a(\mathbf{x},\mathbf{z},\lambda)
\end{equation}
where 
$$H_a(\mathbf{x},\mathbf{z},\lambda)=\frac{p(\mathbf{x}|\mathbf{z}(\mathbf{a}_{-(i,d)}\cup a_{i,d}^*, \mathbf{c}, \lambda))}{p(\mathbf{x}|\mathbf{z}(\mathbf{a}_{-(i,d)}, \mathbf{c}), \lambda))}$$ 

After sampling all the customer links, it samples a table link for all 
customers $(i,d)$ according to
\begin{equation}
\label{cond-c}
p(c_{i,d}^*|\mathbf{a}, \mathbf{c}_{-(i,d)},\mathbf{x}, \lambda)\propto p(c_{i,d}^*)H_c(\mathbf{x},\mathbf{z},\lambda)
\end{equation}
where 
$$H_c(\mathbf{x},\mathbf{z},\lambda)=\frac{p(\mathbf{x}|\mathbf{z}(\mathbf{a}, \mathbf{c}_{-(i,d)}\cup c_{i,d}^*, \lambda))}{p(\mathbf{x}|\mathbf{z}(\mathbf{a}, \mathbf{c}_{-(i,d)}), \lambda))}$$ 

For those customers $(i,d)$ that did not start a new table, i.e.\ with $a_{i,d}\ne (i,d)$, the table link $c_{i,d}^*$ does not affect the clustering, and so 
$H_c(\mathbf{x},\mathbf{z},\lambda) = 1$ in this case.

\begin{figure}[ht]
\centering
\includegraphics[width=0.8\linewidth]{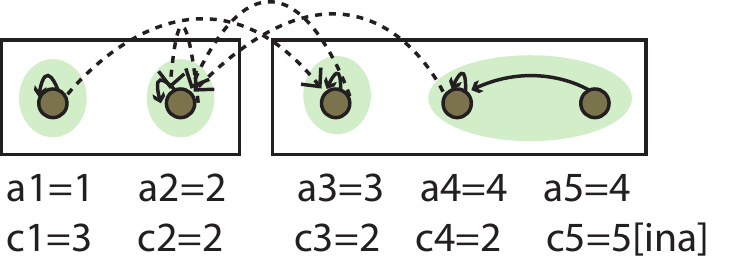}
\caption{\label{fig:mention_example}\small An example of event clustering and the corresponding variable assignments. The assignments of $\mathbf{a}$ induce tables, or within-document (WD) clusters, and the assignments of $\mathbf{c}$ induce menus, or cross-document (CD) clusters. [ina] denotes that the variable is inactive and will not affect the clustering.}
\end{figure}

Referring back to the event coreference example in~\ref{coref_example}, Figure~\ref{fig:mention_example} shows an example of variable configuration for the HDDCRP model and the corresponding coreference clusters. 

In implementation, we can simplify the computations of both $H_a(\mathbf{x},\mathbf{z},\lambda)$ and  $H_c(\mathbf{x},\mathbf{z},\lambda)$ by using the fact that the likelihood under clustering $\mathbf{z}(\mathbf{a},\mathbf{c})$ can be factorized as
$$p(\mathbf{x}|\mathbf{z}(\mathbf{a},\mathbf{c}),\lambda)=\prod_{k\in \mathbf{z}(\mathbf{a},\mathbf{c})}p(\mathbf{x}_{\mathbf{z}=k}|\lambda)$$
where $\mathbf{x}_{\mathbf{z}=k}$ denotes all customers that belong to the global cluster $k$. $p(\mathbf{x}_{\mathbf{z}=k}|\lambda)$ is the marginal probability. It can be computed as
$$p(\mathbf{x}_{\mathbf{z}=k}|\lambda)=
\int p(\phi|\lambda)\prod_{i\in \mathbf{z}=k}p(x_i|\phi)d\phi$$
where $x_i$ is the observation associated with customer $i$. In our problem, the observation corresponds to the lemmatized words in the event mention. We model the observed word counts using cluster-specific multinomial distributions with symmetric Dirichlet priors.

\subsection{Feature-based Distance Functions}
\label{dist-param}
The distance functions $F_{1:D}$ and $F_0$ encode the priors for the clustering distribution, preferring clustering data points that are closer to each other. We consider event mentions as the data points and encode the similarity (or compatibility) between event mentions as priors for event clustering. Specifically, we use a log-linear model to estimate the similarity between a pair of event mentions $(x_i,x_j)$
\begin{equation}
\label{pairmodel}
f_{\boldsymbol\theta}(x_i,x_j)\propto\exp\{\boldsymbol\theta^T \psi(x_i,x_j)\}
\end{equation}
where $\psi$ is a feature vector, containing a rich set of features based on event mentions $i$ and $j$: (1) head word string match, (2) head POS pair, (3) cosine similarity between the head word embeddings (we use the pre-trained 300-dimensional word embeddings from word2vec\footnote{\url{https://code.google.com/p/word2vec/}}), (4) similarity between the words in the event mentions (based on term frequency (TF) vectors), (5) the Jaccard coefficient between the WordNet synonyms of the head words, and (6) similarity between the context words (a window of three words before and after each event mention). If both event mentions involve participants, we consider the similarity between the words in the participant mentions based on the TF vectors, similarly for the time mentions and the location mentions. If the SRL role information is available, we also consider the similarity between words in each SRL role, i.e.\ Arg0, Arg1, Arg2. 

\textbf{Training} We train the parameter $\boldsymbol\theta$ using logistic regression with an L2 regularizer. We construct the training data by considering all ordered pairs of event mentions within a document, and also all pairs of event mentions across similar documents. To measure document similarity, we collect all mentions of events, participants, times and locations in each document and compute the cosine similarity between the TF vectors constructed from all the event-related mentions. We consider two documents to be similar if their TF-based similarity is above a threshold $\sigma$ (we set it to $0.4$ in our experiments).

After learning $\boldsymbol\theta$, we set the within-document distances as $F_d(i,j)=f_{\boldsymbol\theta}(x_i, x_j)$, and the across-document distances as $F_0((i,d),(j,d'))=w(d, d')f_{\boldsymbol\theta}(x_{i,d}, x_{j,d'})$, where $w(d, d')=\exp(\gamma sim(d,d'))$ captures document similarity where $sim(d, d')$ is the TF-based similarity between document $d$ and $d'$, and $\gamma$ is a weight parameter. Higher $\gamma$ leads to a higher effect of document-level similarities on the linking probabilities. We set $\gamma=1$ in our experiments. 

\section{Experiments}
\label{exp}
We conduct experiments using the ECB+ corpus~\cite{cybulska2014using}, the largest available dataset with annotations of both within-document (WD) and cross-document (CD) event coreference resolution. It extends ECB 0.1~\cite{lee2012joint} and ECB~\cite{bejan2010unsupervised} by adding event argument and argument type annotations as well as adding more news documents. The cross-document coreference annotations only exist in documents that describe the same seminal event (the event that triggers the topic of the document and has interconnections with the majority of events from its surrounding textual context~\cite{bejan2014unsupervised}). We divide the dataset into a training set (topics 1-20), a development set (topics 21-23), and a test set (topics 24-43). Table~\ref{data} shows the statistics of the data.

\begin{table}
\begin{footnotesize}
\begin{center}
\scalebox{0.9}{
\begin{tabular}{c|c|c|c|c}
\hline
 & Train & Dev & Test & Total\\
\hline
\# Documents & 462 & 73 & 447 & 982\\
\hline
\# Sentences & 7,294 & 649 & 7,867 & 15,810\\
\hline
\# Annotated event mentions & 3,555 & 441 & 3,290 & 7,286\\
\hline
\# Cross-document chains & 687 & 47 & 486 & 1,220\\
\hline
\# Within-document chains & 2,499 & 316 & 2,137 & 4,952\\
\hline
\end{tabular}
}
\end{center}
\caption{\label{data}Statistics of the ECB+ corpus}
\end{footnotesize}
\end{table}

We performed event coreference resolution on all possible event mentions that are expressed in the documents. Using the event extraction method described in Section~\ref{event_extract}, we extracted 53,429 event mentions, 43,682 participant mentions, 5,791 time mentions and 3,836 location mentions in the test data, covering 93.5\%, 89.0\%, 95.0\%, 72.8\% of the annotated event mentions, participants, time and locations, respectively.

We evaluate both within- and cross-document event coreference resolution. As in previous work~\cite{bejan2010unsupervised}, we evaluate cross-document coreference resolution by merging all documents from the same seminal event into a meta-document and then evaluate the meta-document as in within-document coreference resolution. However, during inference time, we do not assume the knowledge of the mapping of documents to seminal events.

We consider three widely used coreference resolution metrics: (1) MUC~\cite{vilain1995model}, which measures how many gold (predicted) cluster merging operations are needed to recover each predicted (gold) cluster; (2) B$^3$~\cite{bagga1998algorithms}, which measures the proportion of overlap between the predicted and gold clusters for each mention and computes the average scores; and (3) CEAF~\cite{luo2005coreference} (CEAF$_e$), which measures the best alignment of the gold-standard and predicted clusters. We also consider the CoNLL F1, which is the average F1 of the above three measures. All the scores are computed using the latest version (v8.01) of the official CoNLL scorer~\cite{pradhan2014scoring}.

\subsection{Baselines}
We compare our proposed HDDCRP model ({\sc hddcrp}) to five baselines: 
\begin{itemize}
\item {\sc Lemma}: a heuristic method that groups all event mentions, either within or across documents, which have the same lemmatized head word. It is usually considered a strong baseline for event coreference resolution.
\item {\sc Agglomerative}: a supervised clustering method for within-document event coreference~\cite{chen2009pairwise}. We extend it to within- and cross-document event coreference by performing single-link clustering in two phases: first grouping mentions within documents and then grouping within-document clusters to larger clusters across documents. We compute the pairwise-linkage scores using the log-linear model described in Section~\ref{dist-param}.
\item {\sc hdp-lex}: an unsupervised Bayesian clustering model for within- and cross-document event coreference~\cite{bejan2010unsupervised}\footnote{We re-implement the proposed HDP-based models: the {\sc HDP}$_{1f}$, {\sc HDP}$_{flat}$ (including {\sc HDP}$_{flat}$ (LF), (LF+WF), and (LF+WF+SF)) and {\sc HDP}$_{struct}$, but found that the {\sc HDP}$_{flat}$ with lexical features (LF) performs the best in our experiments. We refer to it as {\sc hdp-lex}.}. It is a hierarchical Dirichlet process (HDP) model with the likelihood of all the lemmatized words observed in the event mentions. In general, the HDP can be formulated using a two-level sequential CRP. Our {\sc hddcrp} model is a two-level DDCRP that generalizes the HDP to allow data dependencies to be incorporated at both levels\footnote{Note that {\sc hdp-lex} is not a special case of {\sc hddcrp} because we define the table-level distance function as the distances between customers instead of between tables. In our model, the probability of linking a table $t$ to another table $s$ depends on the distance between the head customer at table $t$ and all other customers who sit at table $s$. Defining the table-level distance function this way allows us to derive a tractable inference algorithm using Gibbs sampling.}. 
\item {\sc ddcrp}: a DDCRP model we develop for event coreference resolution. It applies the distance prior in Equation~\ref{DDCRP-prior} to all pairs of event mentions in the corpus, ignoring the document boundaries. It uses the same likelihood function and the same log-linear model to learn the distance values as {\sc hddcrp}. But it has fewer link variables than  {\sc hddcrp} and it does not distinguish between the within-document and cross-document link variables. For the same clustering structure, {\sc hddcrp} can generate more possible link configurations than {\sc ddcrp}.
\item {\sc hddcrp}$^*$: a variant of the proposed {\sc hddcrp} that only incorporates the within-document dependencies but not the cross-document dependencies. The generative process of {\sc hddcrp}$^*$ is similar to the one described in Section~\ref{sec-hddcrp}, except that in step 2, for each table $t$, we sample a cluster assignment $c_t$ according to 
\begin{equation*}
    p(c_t=k)\propto 
\begin{cases}
    n_k, & k \leq K\\
    \alpha_0, & k = K+1\\
\end{cases}
\end{equation*}
where $K$ is the number of existing clusters, $n_k$ is the number of existing tables that belong to cluster $k$, $\alpha$ is the concentration parameter. And in step 3, the clusters $\mathbf{z}(\mathbf{a}, \mathbf{c})$ are constructed by traversing the customer links and looking up the cluster assignments for the obtained tables. We also use Gibbs sampling for inference. 
\end{itemize} 

\begin{table*}[bth]
\begin{center}
\scalebox{0.9}{
\begin{tabular}{c|c|c|c|c|c|c|c|c|c|c}
\hline
& \multicolumn{3}{c|}{MUC} &\multicolumn{3}{c|}{$B^3$} & \multicolumn{3}{c|}{CEAF$_e$} & CoNLL\\
 & P & R & F1 & P & R & F1 & P & R & F1 & F1\\
\hline
& \multicolumn{10}{c}{Cross-document Event Coreference Resolution (CD)}\\
\hline
{\sc Lemma} & 75.1 & 55.4 & 63.8 & 71.7 & 39.6 & 51.0 & 36.2 & 61.1 & 45.5 & 53.4\\
\hline
{\sc hdp-lex} & 75.5 & 63.5 & 69.0 & 65.6 & 43.7 & 52.5 & 34.8 & 60.2 & 44.1 & 55.2\\
\hline 
{\sc Agglomerative} & 78.3 & 59.2 & 67.4 & 73.2 & 40.2 & 51.9 & 30.2 & 65.6 & 41.4 & 53.6\\
\hline
{\sc ddcrp} & 79.6 & 58.2 & 67.1 & 78.1 & 39.6 & 52.6 & 31.8 & \bf{69.4} & 43.6 & 54.4\\
\hline
{\sc hddcrp}$^*$ & 77.5 & 66.4 & 71.5 & 69.0 & \bf{48.1} & \bf{56.7} & 38.2 & 63.0 & 47.6 & 58.6\\
\hline
{\sc hddcrp} & \bf{80.3} & \bf{67.1} & \bf{73.1} & \bf{78.5} & 40.6 & 53.5 & \bf{38.6} & 68.9 & \bf{49.5} & \bf{58.7}\\
\hline
\hline
& \multicolumn{10}{c}{Within-document Event Coreference Resolution (WD)}\\
\hline
{\sc Lemma} & 60.9 & 30.2 & 40.4 & 78.9 & 57.3 & 66.4 & 63.6 & 69.0 & 66.2 & 57.7\\
\hline
{\sc hdp-lex}  & 50.0 & 39.1 & 43.9 & 74.7 & 67.6 & 71.0 & 66.2 & 71.4 & 68.7 & 61.2\\
\hline
{\sc Agglomerative}& 61.9 & 39.2 & 48.0 & 80.7 & 67.6 & 73.5 & 65.6 & 76.0 & 70.4 & 63.9 \\
\hline
{\sc ddcrp} & 71.2 & 36.4 & 48.2 & 85.4 & 64.9 & 73.8 & 61.8 & 76.1 & 68.2 & 63.4\\
\hline
{\sc hddcrp}$^*$ & 58.1 & \bf{42.8} & 49.3 & 78.4 & \bf{68.7} & 73.2 & \bf{67.6} & 74.5 & 70.9 & 64.5\\
\hline
{\sc hddcrp} & \bf{74.3} & 41.7 & \bf{53.4} & \bf{85.6} & 67.3 & \bf{75.4} & 65.1 & \bf{79.8} & \bf{71.7} & \bf{66.8}\\
\hline
\end{tabular}
}
\caption{\label{main_results} Within- and cross-document coreference results on the ECB+ corpus}
\end{center}
\end{table*}

\subsection{Parameter settings}
For all the Bayesian models, the reported results are averaged results over five MCMC runs, each for $500$ iterations. We found that mixing happens before $500$ iterations in all models by observing the joint log-likelihood. For the {\sc ddcrp}, {\sc hddcrp}$^*$ and {\sc hddcrp}, we randomly initialized the link variables. Before initialization, we assume that each mention belongs to its own cluster. We assume mentions are ordered according to their appearance within a document, but we do not assume any particular ordering of documents. We also truncated the pairwise mention similarity to zero if it is below $0.5$ as we found that it leads to better performance on the development set. We set $\alpha_1=...=\alpha_D=0.5$, $\alpha_0=0.001$ for {\sc hddcrp}, $\alpha_0=1$ for {\sc hddcrp}$^*$, $\alpha=0.1$ for {\sc ddcrp}, and $\lambda=10^{-7}$. All the hyperparameters were set based on the development data.

\subsection{Main Results}
Table~\ref{main_results} shows the event coreference results. We can see that {\sc Lemma}-matching is a strong baseline for event coreference resolution. {\sc hdp-lex} provides noticeable improvements, suggesting the benefit of using an infinite mixture model for event clustering. {\sc Agglomerative} further improves the performance over {\sc hdp-lex} for WD resolution, however, it fails to improve CD resolution. We conjecture that this is due to the combination of ineffective thresholding and the prediction errors on the pairwise distances between mention pairs across documents. Overall,  {\sc hddcrp}$^*$ outperforms all the baselines in CoNLL F1 for both WD and CD evaluation. The clear performance gains over {\sc hdp-lex} demonstrate that it is important to account for pairwise mention dependencies in the generative modeling of event clustering. The improvements over {\sc Agglomerative} indicate that it is more effective to model mention-pair dependencies as clustering priors than as heuristics for deterministic clustering. 

Comparing among the HDDCRP-related models, we can see that {\sc hddcrp} clearly outperforms {\sc ddcrp}, demonstrating the benefits of incorporating the hierarchy into the model. {\sc hddcrp} also performs better than {\sc hddcrp}$^*$ in WD CoNLL F1, indicating that incorporating cross-document information helps within-document clustering. We can also see that {\sc hddcrp} performs similarly to {\sc hddcrp}$^*$ in CD CoNLL F1 due to the lower B$^3$ F1, in particular, the decrease in B$^3$ recall. This is because applying the DDCRP prior at both within- and cross-document levels results in more conservative clustering and produces smaller clusters. This could be potentially improved by employing more accurate similarity priors.    

To further understand the effect of modeling mention-pair dependencies, we analyze the impact of the features in the mention-pair similarity model. Table~\ref{feature_analysis} lists the learned weights of some top features (sorted by weights). We can see that they mainly serve to discriminate event mentions based on the head word similarity (especially embedding-based similarity) and the context word similarity. Event argument information such as \textit{SRL Arg1}, \textit{SRL Arg0}, and \textit{Participant} are also indicative of the coreferential relations. 

\subsection{Discussion}
We found that {\sc hddcrp} corrects many errors made by the traditional agglomerative clustering model ({\sc Agglomerative}) and the unsupervised generative model ({\sc hdp-lex}).  {\sc Agglomerative} easily suffers from error propagation as the errors made by the supervised distance learner cannot be corrected.  {\sc hdp-lex} often mistakenly groups mentions together based on word co-occurrence statistics but not the apparent similarity features in the mentions. In contrast, {\sc hddcrp} avoids such errors by performing probabilistic modeling of clustering and making use of rich linguistic features trained on available annotated data. For example, {\sc hddcrp} correctly groups the event mention ``unveiled" in \textit{``Apple's Phil Schiller \underline{unveiled} a revamped MacBook Pro today"} together with the event mention ``announced" in \textit{``this notebook isn't the only laptop Apple \underline{announced} for the MacBook Pro lineup today"}, while both {\sc hdp-lex} and {\sc Agglomerative} models fail to make such connection. 

By looking further into the errors, we found that a lot of mistakes made by {\sc hddcrp} are due to the errors in event extraction and pairwise linkage prediction. The event extraction errors include false positive and false negative event mentions and event arguments, boundary errors for the extracted mentions, and argument association errors. The pairwise linking errors often come from the lack of semantic and world knowledge, and this applies to both event mentions and event arguments, especially for time and location arguments which are less likely to be repeatedly mentioned and in many cases require external knowledge to resolve their meanings, e.g., ``\textit{May 3, 2013}" is ``\textit{Friday}" and ``\textit{Mount Cook}" is ``\textit{New Zealand's highest peak}". 

\begin{table}
\begin{footnotesize}
\begin{center}
\begin{tabular}{c|c}
\hline
\textbf{Features} & \textbf{Weight}\\
\hline
Head Embedding sim & 4.5 \\
\hline
String match & 2.77\\
\hline
Context sim & 1.75\\
\hline
Synonym sim & 1.56\\
\hline
TF sim & 1.17\\
\hline
SRL Arg1 sim& 1.10\\
\hline
SRL Arg0 sim & 0.89\\
\hline
Participant sim& 0.68\\
\hline
\end{tabular}
\end{center}
\caption{\label{feature_analysis}Learned weights for selected features}
\end{footnotesize}
\end{table}

\section{Conclusion}
In this paper we propose a novel Bayesian model for within- and cross-document event coreference resolution. It leverages the advantages of generative modeling of coreference resolution and feature-rich discriminative modeling of mention reference relations. We have shown its power in resolving event coreference by comparing it to a traditional agglomerative clustering approach and a state-of-the-art unsupervised generative clustering approach. It is worth noting that our model is general and can be easily applied to other clustering problems involving feature-rich objects and cluster sharing across data groups. While the model can effectively cluster objects of a single type, it would be interesting to extend it to allow joint clustering of objects of different types, e.g., events and entities.


\section*{Acknowledgments}
We thank Cristian Danescu-Niculescu-Mizil, Igor Labutov, Lillian Lee, Moontae Lee, Jon Park, Chenhao Tan, and other Cornell NLP seminar participants and the reviewers for their helpful comments. This work was supported in part by NSF grant IIS-1314778 and DARPA DEFT Grant FA8750-13-2-0015. The third author was supported by NSF CAREER CMMI-1254298, NSF IIS-1247696, AFOSR FA9550-12-1-0200, AFOSR FA9550-15-1-0038, and the ACSF AVF. The views and conclusions contained herein are those of the authors and should not be interpreted as necessarily representing the official policies or endorsements, either expressed or implied, of NSF, DARPA or the U.S.\ Government.

\bibliographystyle{acl2012}
\bibliography{acl2015}

\end{document}